\def\year{2020}\relax
\documentclass[letterpaper]{article} 
\usepackage{aaai20}  
\usepackage{times}  
\usepackage{helvet} 
\usepackage{courier}  
\usepackage[hyphens]{url}  
\usepackage{graphicx} 
\usepackage{graphicx}  
\usepackage{amssymb}
\usepackage{amsmath}
\usepackage{multirow}
\usepackage{arydshln}

\title{Message Passing Attention Networks for Document Understanding}
\author{Giannis Nikolentzos,\textsuperscript{\rm 1} Antoine J.-P. Tixier,\textsuperscript{\rm 1} Michalis Vazirgiannis\textsuperscript{\rm 1,2}\\ 
\textsuperscript{\rm 1}\'Ecole Polytechnique\\ 
\textsuperscript{\rm 2}Athens University of Economics and Business\\ 
\{nikolentzos,anti5662,mvazirg\}@lix.polytechnique.fr 
}
 
\begin{document}

\maketitle

\begin{abstract}
Graph neural networks have recently emerged as a very effective framework for processing graph-structured data. These models have achieved state-of-the-art performance in many tasks. Most graph neural networks can be described in terms of message passing, vertex update, and readout functions. In this paper, we represent documents as word co-occurrence networks and propose an application of the message passing framework to NLP, the Message Passing Attention network for Document understanding (MPAD). We also propose several hierarchical variants of MPAD. Experiments conducted on 10 standard text classification datasets show that our architectures are competitive with the state-of-the-art. Ablation studies reveal further insights about the impact of the different components on performance. Code is publicly available at: \url{https://github.com/giannisnik/mpad}.
\end{abstract}

\section{Introduction}
The concept of message passing over graphs has been around for many years \cite{weisfeiler1968reduction,murphy1999loopy}, as well as that of graph neural networks (GNNs) \cite{gori2005new,scarselli2008graph}. However, GNNs have only recently started to be closely investigated, following the advent of deep learning. Some notable examples include \cite{duvenaud2015convolutional,battaglia2016interaction,li2015gated,defferrard2016convolutional,kearnes2016molecular,kipf2016semi,hamilton2017inductive,velivckovic2017graph,xu2018representation}. These approaches are known as \textit{spectral}. Their similarity with message passing (MP) was observed by \cite{kipf2016semi} and formalized by \cite{gilmer2017neural} and \cite{xu2018powerful}. 

The MP framework is based on the core idea of \textit{recursive neighborhood aggregation}. That is, at every iteration, the representation of each vertex is updated based on messages received from its neighbors. The majority of the spectral GNNs can be described in terms of the MP framework.

GNNs have been applied with great success to bioinformatics and social network data, for node classification, link prediction, and graph classification. However, a few studies only have focused on the application of the MP framework to representation learning on text. This paper proposes one such application. More precisely, we represent documents as word co-occurrence networks, and develop an expressive MP GNN tailored to document understanding, the Message Passing Attention network for Document understanding (MPAD). We also propose several hierarchical variants of MPAD. Evaluation on 10 document classification datasets shows that our architectures learn representations that are competitive with the state-of-the-art. Furthermore, ablation experiments shed light on the impact of various architectural choices.

In what follows, we first provide some background about the MP framework (sec.~\ref{sec:mp_framework}), thoroughly describe and explain MPAD (sec.~\ref{sec:mpad}), present our experimental framework (sec.~\ref{sec:exps}), report and interpret our results (sec.~\ref{sec:results}), and provide a review of the relevant literature (sec.~\ref{sec:related}).

\section{Message Passing Neural Networks}\label{sec:mp_framework}
\cite{gilmer2017neural} proposed a MP framework under which many of the recently introduced GNNs can be reformulated\footnote{Note that some GNNs, known as \textit{spatial}, are not based on MP \cite{niepert2016learning,nikolentzos2018kernel,tixier2019graph}.}. MP consists in an aggregation phase followed by a combination phase \cite{xu2018powerful}. More precisely, let $G=(V,E)$ be a graph, and let us consider $v \in V$. At time $t+1$, a message vector $\mathbf{m}_v^{t+1}$ is computed from the representations of the neighbors $\mathcal{N}(v)$ of $v$:
\begin{equation}
    \mathbf{m}_v^{t+1} = {\small \textsc{AGGREGATE}}^{t+1}\big(\big\{\mathbf{h}_w^t \mid w \in \mathcal{N}(v) \big\}\big)
\end{equation}
The new representation $\mathbf{h}^{t+1}_v$ of $v$ is then computed by combining its current feature vector $\mathbf{h}^{t}_v$ with the message vector $\mathbf{m}_v^{t+1}$:
\begin{equation}
    \mathbf{h}_v^{t+1} = {\small \textsc{COMBINE}}^{t+1}\big(\mathbf{h}_v^t, \mathbf{m}_v^{t+1}\big)
\end{equation}
Messages are passed for $T$ time steps. Each step is implemented by a different layer of the MP network. Hence, iterations correspond to network depth. The final feature vector $\mathbf{h}_v^T$ of $v$ is based on messages propagated from all the nodes in the subtree of height $T$ rooted at $v$. It captures both the topology of the neighborhood of $v$ and the distribution of the vertex representations in it.

If a graph-level feature vector is needed, e.g., for classification or regression, a {\small \textsc{READOUT}} pooling function, that must be invariant to permutations, is applied:
\begin{equation}
    \mathbf{h}_G = {\small \textsc{READOUT}}\big(\big\{\mathbf{h}_v^T \mid v \in V \big\}\big)
\end{equation}
Next, we present the MP network we developed for document understanding.

\section{Message Passing Attention network for Document understanding (MPAD)}\label{sec:mpad}

\subsection{Word co-occurrence networks}\label{sub:gow}
We represent a document as a statistical word co-occurrence network \cite{mihalcea2004textrank} with a sliding window of size 2 overspanning sentences. Let us denote that graph by $G=(V,E)$. Each unique word in the preprocessed document is represented by a node in $G$, and an edge is added between two nodes if they are found together in at least one instantiation of the window. $G$ is directed and weighted: edge directions and weights respectively capture text flow and co-occurrence counts.

$G$ is a compact representation of its document. In $G$, immediate neighbors are consecutive words in the same sentence\footnote{except for words at the end/beginning of two successive sentences.}. That is, paths of length 2 correspond to bigrams. Paths of length more than 2 can correspond either to traditional $n$-grams or to \textit{relaxed} $n$-grams, that is, words that never appear in the same sentence but co-occur with the same word(s). Such nodes are linked through common neighbors.

\noindent \textbf{Master node}. Inspired by \cite{scarselli2008graph}, our graph $G$ also includes a special document node, linked to all other nodes via unit weight bi-directional edges. In what follows, let us denote by $n$ the number of nodes in $G$, including the master node.

\subsection{Message passing}\label{sub:mp}
We formulate our {\small \textsc{AGGREGATE}} function as:
\begin{equation}\label{eq:agg}
    \mathbf{M}^{t+1} = {\small \textsc{MLP}}^{t+1}\big(\mathbf{D}^{-1}\mathbf{A}\mathbf{H}^{t}\big) 
\end{equation}
where $\mathbf{H}^t \in \mathbb{R}^{n \times d}$ contains node features ($d$ is a hyperparameter\footnote{at $t$=0, $d$ is equal to the dimensionality of the pretrained word embeddings.}), and $\mathbf{A} \in \mathbb{R}^{n \times n}$ is the adjacency matrix of $G$. Since $G$ is directed, $\mathbf{A}$ is asymmetric. Also, $\mathbf{A}$ has zero diagonal as we choose not to consider the feature of the node itself, only that of its incoming neighbors, when updating its representation\footnote{the feature of the node itself is already taken into account by our GRU-based {\small \textsc{COMBINE}} function (see Eq.~\ref{eq:gru_combine}).}.
Since $G$ is weighted, the $i^{th}$ row of $A$ contains the weights of the edges incoming on node $v_i$. $\mathbf{D} \in \mathbb{R}^{n \times n}$ is the diagonal in-degree matrix of $G$. {\small \textsc{MLP}} denotes a multi-layer perceptron, and $\mathbf{M}^{t+1} \in \mathbb{R}^{n \times d}$ is the message matrix. 

The use of a {\small \textsc{MLP}} was motivated by the observation that for graph classification, MP neural nets with 1-layer perceptrons are inferior to their {\small \textsc{MLP}} counterparts \cite{xu2018powerful}. Indeed, 1-layer perceptrons are not universal approximators of multiset functions. Note that like in \cite{xu2018powerful}, we use a different {\small \textsc{MLP}} at each layer.

\noindent \textbf{Renormalization}.
The rows of $\mathbf{D}^{-1}\mathbf{A}$ sum to 1. This is equivalent to the renormalization trick of \cite{kipf2016semi}, but using only the in-degrees. 
That is, instead of computing a weighted sum of the incoming neighbors' feature vectors, we compute a weighted average of them. The coefficients are proportional to the strength of co-occurrence between words. One should note that by averaging, we lose the ability to distinguish between different neighborhood structures in some special cases, that is, we lose \textit{injectivity}. Such cases include neighborhoods in which all nodes have the same representations, and neighborhoods of different sizes containing various representations in equal proportions \cite{xu2018powerful}. As suggested by the results of an ablation experiment, averaging is better than summing in our application (see subsection~\ref{sub:ablation}). Note that instead of simply summing/averaging, we also tried using GAT-like attention \cite{velivckovic2017graph} in early experiments, without obtaining better results.

As far as our {\small \textsc{COMBINE}} function, we use the Gated Recurrent Unit \cite{cho2014learning,chung2014empirical}:
\begin{align}\label{eq:gru_combine}
\mathbf{H}^{t+1} & = {\small \textsc{GRU}} (\mathbf{H}^{t}, \mathbf{M}^{t+1})
\end{align}
Omitting biases for readability, we have:
\begin{equation}
\label{eq:gru}
\begin{split}
    \mathbf{R}^{t+1} &= \sigma(\mathbf{W}_R^{t+1} \mathbf{M}^{t+1} + \mathbf{U}_R^{t+1} \mathbf{H}^t)\\
  \mathbf{Z}^{t+1} &= \sigma(\mathbf{W}_Z^{t+1} \mathbf{M}^{t+1} + \mathbf{U}_Z^{t+1} \mathbf{H}^t)\\
  \tilde{\mathbf{H}}^{t+1} &= \mathrm{tanh}(\mathbf{W}^{t+1} \mathbf{M}^{t+1} + \mathbf{U}^{t+1}(\mathbf{R}^{t+1} \odot \mathbf{H}^t))\\
  \mathbf{H}^{t+1} &= (1 - \mathbf{Z}^{t+1}) \odot \mathbf{H}^t + \mathbf{Z}^{t+1} \odot \tilde{\mathbf{H}}^{t+1}
\end{split}
\end{equation}
where the $\mathbf{W}$ and $\mathbf{U}$ are trainable weight matrices not shared across time steps, $\sigma(\mathbf{x}) = 1/(1+\exp(-\mathbf{x}))$ is the sigmoid function, and $\mathbf{R}$ and $\mathbf{Z}$ are the parameters of the reset and update gates. The reset gate controls the amount of information from the previous time step (in $\mathbf{H}^t$) that should propagate to the candidate representations, $\tilde{\mathbf{H}}^{t+1}$. The new representations $\mathbf{H}^{t+1}$ are finally obtained by linearly interpolating between the previous and the candidate ones, using the coefficients returned by the update gate.

\noindent \textbf{Interpretation}. Updating node representations through a GRU should in principle allow nodes to encode a combination of local and global signals (low and high values of $t$, resp.), by allowing them to remember about past iterations. In addition, we also explicitly consider node representations at all iterations when reading out (see Eq.~\ref{eq:output}).

\subsection{Readout}\label{sub:readout}
After passing messages and performing updates for $T$ iterations, we obtain a matrix $\mathbf{H}^T \in \mathbb{R}^{n \times d}$ containing the final vertex representations. Let $\hat{G}$ be graph $G$ without the special document node and its adjacent edges, and matrix $\mathbf{\hat{H}}^T \in \mathbb{R}^{(n-1) \times d}$ be the corresponding representation matrix (i.e., $\mathbf{H}^T$ without the row of the document node).

We use as our {\small \textsc{READOUT}} function the concatenation of self-attention applied to $\mathbf{\hat{H}}^T$ with the final document node representation. More precisely, we apply a global self-attention mechanism \cite{lin2017structured} to the rows of $\mathbf{\hat{H}}^T$. As shown in Eq.~\ref{eq:att}, $\mathbf{\hat{H}}^T$ is first passed to a dense layer parameterized by matrix $\mathbf{W}_A^T \in \mathbb{R}^{d \times d}$. An alignment vector $\boldsymbol{\alpha}$ is then derived by comparing, via dot products, the rows of the output of the dense layer $\mathbf{Y}^T \in \mathbb{R}^{(n-1) \times d}$ with a trainable vector $\mathbf{v}^T \in \mathbb{R}^d$ (initialized randomly) and normalizing with a softmax. The normalized alignment coefficients are finally used to compute the attentional vector $\mathbf{u}^T \in \mathbb{R}^d$ as a weighted sum of the final representations $\mathbf{\hat{H}}^T$.
\begin{equation}\label{eq:att}
  \begin{split}
    \mathbf{Y}^T &= \mathrm{tanh}( \mathbf{\hat{H}}^T \mathbf{W}^T_A)\\
    \boldsymbol{\alpha}^T_i &= \frac{\exp(\mathbf{Y_i}^T\cdot \mathbf{v}^T)}{\sum_{j=1}^{n-1} \exp(\mathbf{Y_j}^T\cdot \mathbf{v}^T)}\\
    \mathbf{u}^T &= \sum_{i=1}^{n-1} \boldsymbol{\alpha}^T_i \mathbf{\hat{H}}^{T}_{i}
  \end{split}
\end{equation}
Note that we tried with multiple context vectors, i.e., with a matrix $\mathbf{V}^T$ instead of a vector $\mathbf{v}^T$, like in \cite{lin2017structured}, but results were not convincing, even when adding a regularization term to the loss to favor diversity among the rows of $\mathbf{V}^T$.

\noindent \textbf{Master node skip connection}. $\mathbf{h}_G^T \in \mathbb{R}^{2d}$ is obtained by concatenating $\mathbf{u}^T$ and the final master node representation. That is, the master node vector bypasses the attention mechanism. This is equivalent to a skip or shortcut connection \cite{he2016deep}. The reason behind this choice is that we expect the special document node to learn a high-level summary about the document, such as its size, vocabulary, etc. (more details are given in subsection~\ref{sub:ablation}). Therefore, by making the master node bypass the attention layer, we directly inject global information about the document into its final representation.

\noindent \textbf{Multi-readout}. \cite{xu2018powerful}, inspired by Jumping Knowledge Networks \cite{xu2018representation}, recommend to not only use the final representations when performing readout, but also that of the earlier steps. Indeed, as one iterates, node features capture more and more global information. However, retaining more local, intermediary information might be useful too. Thus, instead of applying the readout function only to $t=T$, we apply it to all time steps and concatenate the results, finally obtaining $\mathbf{h}_G \in \mathbb{R}^{T \times 2d}$~:
\begin{equation}\label{eq:output}
    \mathbf{h}_G = {\small\textsc{CONCAT}} \big( {\small \textsc{READOUT}} \big(\mathbf{H}^t\big)\mid t=1 \dots T\big) 
\end{equation}
In effect, with this modification, we take into account features based on information aggregated from subtrees of different heights (from 1 to $T$), corresponding to local and global features.

\subsection{Hierarchical variants of MPAD}
Through the successive MP iterations, it could be argued that MPAD implicitly captures some soft notion of the hierarchical structure of documents (words $\rightarrow$ bigrams $\rightarrow$ compositions of bigrams, etc.). However, it might be beneficial to explicitly capture document hierarchy. Hierarchical architectures have brought significant improvements to many NLP tasks, such as language modeling and generation \cite{lin2015hierarchical,li2015hierarchical}, sentiment and topic classification \cite{tang2015document,yang2016hierarchical}, and spoken language understanding \cite{raheja2019dialogue,shang2019energy}. Inspired by this line of research, we propose several hierarchical variants of MPAD, detailed in what follows. In all of them, we represent each sentence in the document as a word co-occurrence network, and obtain an embedding for it by applying MPAD as previously described.

\noindent \textbf{MPAD-sentence-att}.
Here, the sentence embeddings are simply combined through self-attention.

\noindent \textbf{MPAD-clique}.
In this variant, we build a complete graph where each node represents a sentence. We then feed that graph to MPAD, where the feature vectors of the nodes are initialized with the sentence embeddings previously obtained.

\begin{figure}[t]
\centering
\includegraphics[width=0.7\columnwidth]{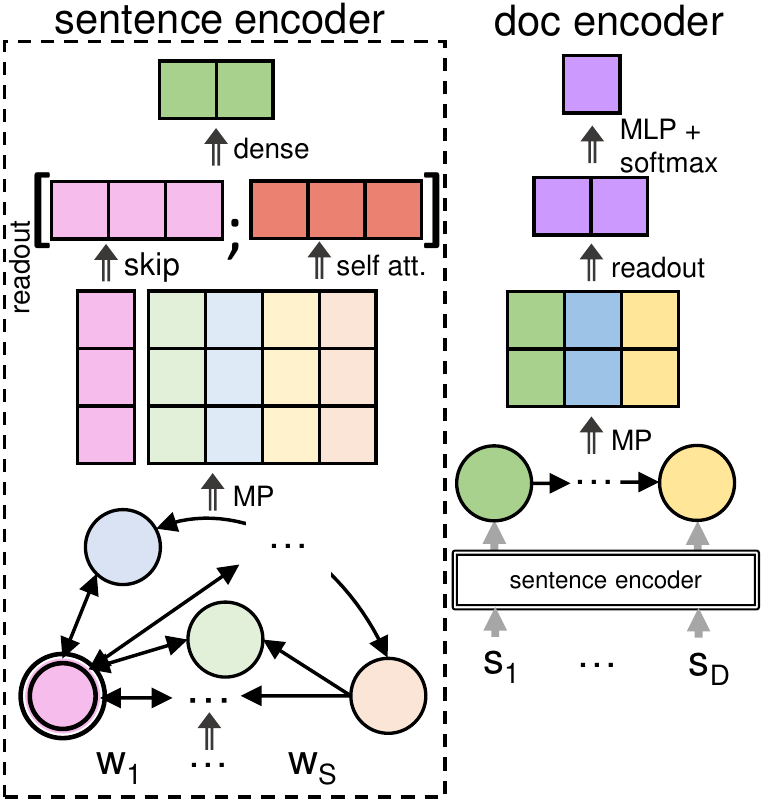}
\caption{Illustration of MPAD-path ($\circledcirc$: master node). \label{fig:architecture}}
\end{figure}

\noindent \textbf{MPAD-path}.
This variant, shown in Fig.~\ref{fig:architecture}, is similar to the clique one, except that instead of a complete graph, we build a path according to the natural flow of the text. That is, two nodes are linked by a directed edge if the two sentences they represent follow each other in the document.

Note that the sentence graphs in MPAD-clique and MPAD-path do not feature a master node.

\section{Experiments}\label{sec:exps}

\subsection{Datasets}
We evaluate the quality of the document embeddings learned by MPAD on 10 document classification datasets, covering the topic identification, coarse and fine sentiment analysis and opinion mining, and subjectivity detection tasks. We briefly introduce the datasets next. Their statistics are reported in Table~\ref{tab:statistics}.

\begin{table*}[t]
\centering
\resizebox{1.5\columnwidth}{!}{
\def\arraystretch{1.1}
\begin{tabular}{cccccccc}
\multirow{2}{*}{\textbf{Dataset}} & \textbf{\# training} & \textbf{\# test} & \multirow{2}{*}{\textbf{\# classes}} & \multirow{2}{*}{\textbf{av. \# words}} & \multirow{2}{*}{\textbf{max \# words}} & \multirow{2}{*}{\textbf{voc. size}} & \textbf{\# pretrained}\\
 & \textbf{examples} & \textbf{examples} & & & & & \textbf{words} \\ \hline
Reuters & 5,485 & 2,189 & 8 & 102.3 & 964 & 23,585 & 15,587 \\ 
Snippets & 10,060 & 2,280 & 8 & 18.0 & 50 & 29,257 & 17,142 \\ 
BBCSport & 737 & CV & 5 & 380.5 & 1,818 & 14,340 & 13,390 \\ 
Polarity & 10,662 & CV & 2 & 20.3 & 56 & 18,777 & 16,416 \\ 
Subjectivity & 10,000 & CV & 2 & 23.3 & 120 & 21,335 & 17,896 \\ 
MPQA & 10,606 & CV & 2 & 3.0 & 36 & 6,248 & 6,085 \\ 
IMDB & 25,000 & 25,000 & 2 & 254.3 & 2,633 & 141,655 & 104,391 \\ 
TREC & 5,452 & 500 & 6 & 10.0 & 37 & 9,593 & 9,125 \\ 
SST-1 & 157,918 & 2,210 & 5 & 7.4 & 53 & 17,833 & 16,262 \\ 
SST-2 & 77,833 & 1,821 & 2 & 9.5 & 53 & 17,237 & 15,756 \\
Yelp2013 & 301,514 & 33,504 & 5 & 143.7 & 1,184 & 48,212 & 48,212 \\ \hline
\end{tabular}
}
\caption{Statistics of the datasets used in our experiments. CV indicates that cross-validation was used. \# pretrained words refers to the number of words in the vocabulary having an entry in the Google News word vectors (except for Yelp2013).}
\label{tab:statistics}
\end{table*}

\noindent (1) \textbf{Reuters} contains stories from the Reuters news agency. We used the ModApte split, removed documents belonging to multiple classes and considered only the 8 classes with the highest number of training examples.

\noindent (2) \textbf{BBCSport} \cite{greene2006practical} contains sports news articles from the BBC Sport website.

\noindent (3) \textbf{Polarity} \cite{pang2005seeing} features positive and negative labeled snippets from Rotten Tomatoes.

\noindent (4) \textbf{Subjectivity} \cite{pang2004sentimental} contains movie review snippets from Rotten Tomatoes (subjective sentences), and IMDB plot summaries (objective sentences).

\noindent (5) \textbf{MPQA} \cite{wiebe2005annotating} is made of positive and negative phrases, annotated as part of the summer 2002 NRRC Workshop on Multi-Perspective Question Answering.

\noindent (6) \textbf{IMDB} \cite{maas2011learning} is a collection of highly polarized movie reviews (positive/negative).

\noindent (7) \textbf{TREC} \cite{li2002learning} consists of questions that are classified into 6 different categories.

\noindent (8) \textbf{SST-1} \cite{socher2013recursive} contains the same snippets as Polarity, split into multiple sentences and annotated with fine-grained polarity (from very negative to very positive).

\noindent (9) \textbf{SST-2} \cite{socher2013recursive} is the same as SST-1 but with neutral reviews removed and snippets classified as positive or negative.

\noindent (10) \textbf{Yelp2013} \cite{tang2015document} features reviews obtained from the 2013 Yelp Dataset Challenge.

\subsection{Baselines}
We evaluate MPAD against multiple state-of-the-art baseline models, including hierarchical ones, to enable fair comparison with the hierarchical MPAD variants.

\noindent \textbf{Doc2vec} \cite{le2014distributed} is an extension of word2vec that learns vectors for documents in a fully unsupervised manner. Document embeddings are then fed to a logistic regression classifier.

\noindent \textbf{CNN} \cite{kim2014convolutional}. 1D convolutional neural network where the word embeddings are used as channels (depth dimensions).

\noindent \textbf{DAN} \cite{iyyer2015deep}.
The Deep Averaging Network passes the unweighted average of the embeddings of the input words through multiple dense layers and a final softmax.

\noindent \textbf{Tree-LSTM} \cite{tai2015improved} is a generalization of the standard LSTM architecture to constituency and dependency parse trees.

\noindent \textbf{DRNN} \cite{irsoy2014deep}.
Recursive neural networks are stacked and applied to parse trees.

\noindent \textbf{LSTMN} \cite{cheng2016long} is an extension of the LSTM model where the memory cell is replaced by a memory network which stores word representations.

\noindent \textbf{C-LSTM} \cite{zhou2015c} combines convolutional and recurrent neural networks. The region embeddings provided by a CNN are fed to a LSTM.

\noindent \textbf{SPGK} \cite{nikolentzos2017shortest} also models documents as word co-occurrence networks. It computes a graph kernel that compares shortest paths extracted from the word co-occurrence networks and then relies on a SVM.

\noindent \textbf{WMD} \cite{kusner2015word} is an application of the well-known Earth Mover's Distance to text. A $k$-nearest neighbor classifier is used.

\noindent \textbf{DiSAN} \cite{shen2018disan} uses directional self-attention along with multi-dimensional attention to generate document representations.

\noindent \textbf{LSTM-GRNN} \cite{tang2015document} is a hierarchical model where sentence embeddings are obtained with a CNN and a GRU-RNN is fed the sentence representations to obtain a document vector.

\noindent \textbf{HN-ATT} \cite{yang2016hierarchical} is another hierarchical model, where the same encoder architecture (a bidirectional GRU-RNN) is used for both sentences and documents. Self-attention is applied at each level.

\subsection{Model configuration and training}
We preprocess all datasets using the code of \cite{kim2014convolutional}. On Yelp2013, we also replace all tokens appearing strictly less than 6 times with a special \texttt{UNK} token, like in \cite{yang2016hierarchical}. We then build a directed word co-occurrence network from each document, with a window of size 2.

We use two MP iterations ($T$=$2$) for the basic MPAD, and two MP iterations at each level, for the hierarchical variants. The output of the readout goes through a dense layer before reaching the final classification layer (or the next level, at the first level of MPAD-path and MPAD-clique). We set $d$ to $64$, except on IMDB and Yelp on which $d=128$, and use a two-layer {\small \textsc{MLP}}. The final graph representations are passed through a softmax for classification. All our dense layers (except in self-attention) use ReLU activation. We train MPAD in an end-to-end fashion by minimizing the cross-entropy loss function with the Adam optimizer \cite{kingma2014adam} and an initial learning rate of 0.001. 

To regulate potential differences in magnitude, we apply batch normalization after concatenating the feature vector of the master node with the self-attentional vector, that is, after the skip connection (see subsection~\ref{sub:readout}). To prevent overfitting, we use dropout \cite{srivastava2014dropout} with a rate of 0.5. We select the best epoch, capped at 200, based on the validation accuracy. When cross-validation is used (see 3\textsuperscript{rd} column of Table~\ref{tab:statistics}), we construct a validation set by randomly sampling 10\% of the training set of each fold.

On all datasets except Yelp2013, we use the publicly available\footnote{\url{https://code.google.com/archive/p/word2vec}} 300-dimensional pre-trained Google News vectors \cite{mikolov2013distributed} to initialize the node representations $\mathbf{H}^0$. On Yelp2013, we follow \cite{yang2016hierarchical} and learn our own word vectors from the training and validation sets with the gensim implementation of word2vec \cite{rehurek_lrec}. MPAD was implemented in Python 3.6 using the PyTorch library. All experiments were run on a single machine consisting of a 3.4 GHz Intel Core i7 CPU with 16 GB of RAM and an NVidia GeForce Titan Xp GPU.

\section{Results and ablations}\label{sec:results}

\subsection{Results}

\begin{table*}[t]
\centering
\resizebox{2.1\columnwidth}{!}{
\def\arraystretch{1.1}
\begin{tabular}{l|cccccccccc}
\textbf{Model} & Reut. \, & BBC \, & Pol. \, & Subj. \, & MPQA \, & IMDB \, & TREC \, & SST-1 \, & SST-2 \, & Yelp'13 \, \\ \hline
doc2vec \cite{le2014distributed} & 95.34 \, & 98.64 \, & 67.30 \, & 88.27 \, & 82.57 \, & \textbf{92.5} \, \, & 70.80 \, & 48.7 \, \, & 87.8 \, \, & 57.7 \, \, \\
CNN \cite{kim2014convolutional} & 97.21 \, & 98.37 \, & \textbf{81.5} \, \, & 93.4 \, \, & 89.5 \, \, & 90.28 \, & 93.6 \, \, & 48.0 \, \, & 87.2 \, \, & 64.89 \, \\
DAN \cite{iyyer2015deep} & 94.79 \, & 94.30 \, & 80.3 \, \, & 92.44 \, & 88.91 \, & 89.4 \, \, & 89.60 \, & 47.7 \, \, & 86.3 \, \, & 61.55 \, \\ 
Tree-LSTM \cite{tai2015improved} & - \, & - & - \, & - & - \, & - & - \, & 51.0 \, \, & 88.0 \, \, & - \, \\
DRNN \cite{irsoy2014deep} & - \, & - & - \, & - & - \, & - & - \, & 49.8 \, \, & 86.6 \, \, & - \, \\
LSTMN \cite{cheng2016long} & - \, & - & - \, & - & - \, & - & - \, & 47.9 \, \, & 87.0 \, \, & - \, \\
C-LSTM \cite{zhou2015c} & - \, & - & - \, & - & - \, & - & 94.6 \, \, & 49.2 \, \, & 87.8 \, \, & - \, \\
SPGK \cite{nikolentzos2017shortest} & 96.39 \, & 94.97 \, & 77.89 \, & 91.48 \, & 85.78 \, & \texttt{OOM} \, & 90.69 \, & \texttt{OOM} \, & \texttt{OOM} \, & \texttt{OOM} \, \\
WMD \cite{kusner2015word} & 96.5 \, \, & 98.71 \, & 66.42 \, & 86.04 \, & 83.95 \, & \texttt{OOM} \, & 73.40 \, & \texttt{OOM} \, & \texttt{OOM} \, & \texttt{OOM} \, \\
DiSAN \cite{shen2018disan} & 97.35 \, & 96.05 \, & 80.38 \, & \textbf{94.2} \, \, & \textbf{90.1} \, \, & 83.25 \, & 94.2 \, \, & \textbf{51.72} \, & 86.76 \, & 60.51 \, \\ 
LSTM-GRNN \cite{tang2015document} & 96.16 \, & 95.52 \, & 79.98 \, & 92.38 \, & 89.08 \, & 89.98 \, & 89.40 \, & 48.09 \, & 86.38 \, & 65.1 \, \, \\ 
HN-ATT \cite{yang2016hierarchical} & 97.25 \, & 96.73 \, & 80.78 \, & 92.92 \, & 89.08 \, & 90.06 \, & 90.80 \, & 49.00 \, & 86.71 \, & \textbf{68.2} \, \, \\ \hline
MPAD & 97.07 \, & 98.37 \, & 80.24 \, & 93.46* & 90.02 \, & 91.30 \, & \textbf{95.60}* & 49.09 \, & 87.80 \, & 66.16 \, \\
\hdashline
MPAD-sentence-att & 96.89 \, & 99.32 \, & 80.44 \, & 93.02 \, & \textbf{90.12}* & 91.70 \, & \textbf{95.60}* & 49.95* & \textbf{88.30}* & 66.47 \, \\
MPAD-clique  & \textbf{97.57}* & \textbf{99.72}* & 81.17* & 92.82 \, & 89.96 \, & 91.87* & 95.20 \, & 48.86 \, & 87.91 \, & 66.60 \, \\
MPAD-path & 97.44 \, & 99.59 \, & 80.46 \, & 93.31 \, & 89.81 \, & 91.84 \, & 93.80 \, & 49.68 \, & 87.75 \, & 66.80* \\ \hline
\end{tabular}
}
\caption{Classification accuracies. Best performance per column in \textbf{bold}, *best MPAD variant. \texttt{OOM}:~$>$16GB RAM.}
\label{tab:results}
\end{table*}

\begin{table}[t]
\resizebox{\columnwidth}{!}{
\centering
\def\arraystretch{1.1}
\begin{tabular}{l|ccc}
\textbf{MPAD variant} & Reut. & Pol. & IMDB \\ \hline
MPAD 1MP & 96.57 & 79.91 & 90.57 \\
MPAD 2MP* & 97.07 & 80.24 & \textbf{91.30} \\
MPAD 3MP & 97.07 & 80.20 & 91.24 \\
MPAD 4MP & \textbf{97.48} & 80.52 & \textbf{91.30} \\
\hline
MPAD 2MP undirected & 97.35 & 80.05 & 90.97 \\
MPAD 2MP no master node & 96.66 & 79.15 & 91.09 \\ 
MPAD 2MP no renormalization & 96.02 & 79.84 & 91.16 \\ 
MPAD 2MP neighbors-only & 97.12 & 79.22 & 89.50 \\
MPAD 2MP no master node skip connection & 96.93 & \textbf{80.62} & 91.12 \\\hline
\end{tabular}
}
\caption{Ablation results. The $n$ in $n$MP refers to the number of message passing iterations. *vanilla model (MPAD in Table~\ref{tab:results}).}
\label{tab:ablation_results}
\end{table}

Experimental results are shown in Table~\ref{tab:results}. For the baselines, we provide the scores reported in the original papers. Furthermore, we have evaluated some of the baselines on the rest of our benchmark datasets, and we also report these scores. MPAD reaches best performance on 5 out of 10 datasets, and is close second elsewhere. Moreover, the 5 datasets on which MPAD ranks first widely differ in training set size, number of categories, and prediction task (topic, sentiment, etc.), which indicates that MPAD can perform well in different settings.

\noindent \textbf{MPAD vs. hierarchical variants}. On 9 datasets out of 10, one or more of the hierarchical variants outperform the vanilla MPAD architecture, highlighting the benefit of explicitly modeling the hierarchical nature of documents.

However, on Subjectivity, standard MPAD outperforms all hierarchical variants. On TREC, it reaches the same accuracy. We hypothesize that in some cases, using a different graph to separately encode each sentence might be worse than using one single graph to directly encode the document. Indeed, in the single document graph, some words that never appear in the same sentence can be connected through common neighbors, as was explained in subsection~\ref{sub:gow}. So, this way, some notion of cross-sentence context is captured while learning representations of words, bigrams, etc. at each MP iteration. This creates better informed representations, resulting in a better document embedding. With the hierarchical variants, on the other hand, each sentence vector is produced in isolation, without any contextual information about the other sentences in the document. Therefore, the final sentence embeddings might be of lower quality, and as a group might also contain redundant/repeated information. When the sentence vectors are finally combined into a document representation, it is too late to take context into account.

\subsection{Ablation studies}\label{sub:ablation}
To understand the impact of some hyperparameters on performance, we conducted additional experiments on the Reuters, Polarity, and IMDB datasets, with the non-hierarchical version of MPAD. Results are shown in Table~\ref{tab:ablation_results}.

\noindent \textbf{Number of MP iterations}. First, we varied the number of message passing iterations from 1 to 4. We can clearly see in Table~\ref{tab:ablation_results} that having more iterations improves performance. We attribute this to the fact that we are reading out at each iteration from 1 to $T$ (see Eq.~\ref{eq:output}), which enables the final graph representation to encode a mixture of low-level and high-level features. Indeed, in initial experiments involving readout at $t$=$T$ only, setting $T\geq2$ was always decreasing performance, despite the GRU-based updates (Eq.~\ref{eq:gru_combine})\footnote{The GRU should in principle enable nodes to retain locality in their representations, by remembering about early iterations.}. These results were consistent with that of \cite{yao2019graph} and \cite{kipf2016semi}, who both are reading out only at $t$=$T$ too. We hypothesize that node features at $T\geq2$ are too diffuse to be entirely relied upon during readout. More precisely, initially at $t$=$0$, node representations capture information about words, at $t$=$1$, about their 1-hop neighborhood (bigrams), at $t$=$2$, about compositions of bigrams, etc. Thus, pretty quickly, node features become general and diffuse. In such cases, considering also the lower-level, more precise features of the earlier iterations when reading out may be necessary.

\noindent \textbf{Undirected edges}. On Reuters, using an undirected graph leads to better performance, while on Polarity and IMDB, it is the opposite. This can be explained by the fact that Reuters is a topic classification task, for which the presence or absence of some patterns is important, but not necessarily the order in which they appear, while Polarity and IMDB are sentiment analysis tasks. To capture sentiment, modeling word order is crucial, e.g., in detecting negation.

\noindent \textbf{No master node}. Removing the master node deteriorates performance across all datasets, clearly showing the value of having such a node. We hypothesize that since the special document node is connected to all other nodes, it is able to encode during message passing a summary of the document.

\noindent \textbf{No renormalization}. Here, we do not use the renormalization trick of \cite{kipf2016semi} during MP (see subsection~\ref{sub:mp}). That is, Eq.~\ref{eq:agg} becomes $\mathbf{M}^{t+1} = {\small \textsc{MLP}}^{t+1}\big(\mathbf{A}\mathbf{H}^{t}\big)$. In other words, instead of computing a weighted average of the incoming neighbors' feature vectors, we compute a weighted sum of them\footnote{Weights are co-occurrence counts, as before.}. Unlike the mean, which captures distributions, the sum captures structural information \cite{xu2018powerful}. As shown in Table~\ref{tab:ablation_results}, using sum instead of mean decreases performance everywhere, suggesting that in our application, capturing the distribution of neighbor representations is more important that capturing their structure. We hypothesize that this is the case because statistical word co-occurrence networks tend to have similar structural properties, regardless of the topic, polarity, sentiment, etc. of the corresponding documents.

\noindent \textbf{Neighbors-only}. In this experiment, we replaced the GRU {\small \textsc{COMBINE}} function (see Eq.~\ref{eq:gru_combine}) with the identity function. That is, we simply have $\mathbf{H}^{t+1}$=$\mathbf{M}^{t+1}$. Since $\mathbf{A}$ has zero diagonal, by doing so, we completely ignore the previous feature of the node itself when updating its representation. That is, the update is based entirely on its neighbors. Except on Reuters (almost no change), performance always suffers, stressing the need to take into account the root node during updates and not only its neighborhood.

\noindent \textbf{No master node skip connection}. Here, the master node does not bypass the attention mechanism and is treated as a normal node. This leads to better performance on Polarity, but slightly worse performance on Reuters and IMDB.

\section{Related work}\label{sec:related}
\cite{kipf2016semi,atwood2016diffusion,velivckovic2017graph,hamilton2017inductive} conduct some node classification experiments on citation networks, where nodes are scientific papers, i.e., textual data. However, text is only used to derive node feature vectors. The external graph structure, which plays a central role in determining node labels, is completely unrelated to text.

On the other hand, \cite{henaff2015deep,defferrard2016convolutional} experiment on traditional document classification tasks. They both build $k$-nearest neighbor similarity graphs based on the Gaussian diffusion kernel. More precisely, \cite{henaff2015deep} build one single graph where nodes are documents and distance is computed in the BoW space. Node features are then used for classification. Closer to our work, \cite{defferrard2016convolutional} represent each document as a graph. All document graphs are derived from the same underlying structure. Only node features, corresponding to the entries of the documents' BoW vectors, vary. The underlying, shared structure is that of a $k$-NN graph where nodes are vocabulary terms and similarity is the cosine of the word embedding vectors. \cite{defferrard2016convolutional} then perform graph classification. However they found performance to be lower than that of a naive Bayes classifier.

\cite{peng2018large} use a GNN for hierarchical classification into a large taxonomy of topics. This task differs from traditional document classification. The authors represent documents as unweighted, undirected word co-occurrence networks with word embeddings as node features. They then use the \textit{spatial} GNN of \cite{niepert2016learning} to perform graph classification.

The work closest to ours is probably that of \cite{yao2019graph}. The authors adopt the \textit{semi-supervised node classification} approach of \cite{kipf2016semi}. They build one single undirected graph from the entire dataset, with both word and document nodes. Document-word edges are weighted by TF-IDF and word-word edges are weighted by pointwise mutual information derived from co-occurrence within a sliding window. There are no document-document edges. The GNN is trained based on the cross-entropy loss computed only for the labeled nodes, that is, the documents in the training set. When the final node representations are obtained, one can use that of the test documents to classify them and evaluate prediction performance.

There are significant differences between \cite{yao2019graph} and our work. First, our approach is \textit{inductive}\footnote{Note that other GNNs used in inductive settings can be found \cite{hamilton2017inductive,velivckovic2017graph}.}, not \textit{transductive}. Indeed, while the node classification approach of \cite{yao2019graph} requires all test documents at training time, our graph classification model is able to perform inference on new, never-seen documents. The downside of representing documents as separate graphs, however, is that we lose the ability to capture corpus-level dependencies. Also, our directed graphs capture word ordering, which is ignored by \cite{yao2019graph}. Finally, the approach of \cite{yao2019graph} requires computing the PMI for every word pair in the vocabulary, which may be prohibitive on datasets with very large vocabularies. On the other hand, the complexity of MPAD does not depend on vocabulary size.

MPAD is also related to the Transformer's encoder stack \cite{vaswani2017attention}. Specifically, the self-attention layer in each encoder updates the representation of each term based on the representations of all the other terms in the document, and can thus be thought of as a function performing the {\small \textsc{AGGREGATE}} and {\small \textsc{COMBINE}} steps. Stacking multiple encoders can also be thought of as performing multiple MP iterations. The main difference is that the Transformer's self-attention graph is complete, thus ignoring word order and proximity. Also, building that graph requires constructing an adjacency matrix that may become prohibitively large with long documents.

\section{Conclusion}
We proposed an application of the message passing framework to NLP, the Message Passing Attention network for Document understanding (MPAD). Experiments show that our architecture is competitive with the state-of-the-art. By processing weighted, directed word co-occurrence networks, MPAD is sensitive to word order and word-word relationship strength. To capture the hierarchical structure of documents, we also proposed three hierarchical variants of MPAD, that bring improvements over the vanilla model.

\section{Acknowledgments}
GN is supported by the project ``ESIGMA'' (ANR-17-CE40-0028). We thank the NVidia corporation for the donation of a GPU as part of their GPU grant program.

\small
\bibliographystyle{aaai}
\bibliography{biblio}

\end{document}